# Fast Genetic Algorithm for feature selection — A qualitative approximation approach

Mohammed Ghaith Altarabichi [*], Sławomir Nowaczyk, Sepideh Pashami, Peyman Sheikholharam Mashhadi

*Center for Applied Intelligent Systems Research, Halmstad University, Sweden*[1]

A B S T R A C T

Evolutionary Algorithms (EAs) are often challenging to apply in real-world settings since evolutionary computations involve a large number of evaluations of a typically expensive fitness function. For example, an evaluation could involve training a new machine learning model. An approximation (also known as meta-model or a surrogate) of the true function can be used in such applications to alleviate the computation cost. In this paper, we propose a two-stage surrogate-assisted evolutionary approach to address the computational issues arising from using Genetic Algorithm (GA) for feature selection in a wrapper setting for large datasets.

We define "Approximation Usefulness" to capture the necessary conditions to ensure correctness of the EA computations when an approximation is used. Based on this definition, we propose a procedure to construct a lightweight qualitative meta-model by the active selection of data instances. We then use a meta-model to carry out the feature selection task. We apply this procedure to the GA-based algorithm CHC (Cross generational elitist selection, Heterogeneous recombination and Cataclysmic mutation) to create a Qualitative approXimations variant, $CHC_{QX}$. We show that $CHC_{QX}$ converges faster to feature subset solutions of significantly higher accuracy (as compared to CHC), particularly for large datasets with over 100K instances. We also demonstrate the applicability of the thinking behind our approach more broadly to Swarm Intelligence (SI), another branch of the Evolutionary Computation (EC) paradigm with results of $PSO_{QX}$, a qualitative approximation adaptation of the Particle Swarm Optimization (PSO) method. A GitHub repository with the complete implementation is available.[2]

## 1. Introduction

Feature Selection (FS) and Instance Selection (IS) are two well-known data mining techniques used to identify subsets of the most informative features and instances for a given learning task. Feature selection and instance selection primarily aims to achieve two goals: (a) reduce computational complexity by using fewer features, and instances, for model training; (b) improve generalization performance and model accuracy by reducing overfitting. In practice, these tasks are often performed in a greedy manner, since finding the best solution is often intractable, and even meta-heuristic optimizations are prohibitively slow. In this paper, we propose a method that uses active sampling to create an approximate, fast meta-model.

Genetic Algorithm (GA) pioneered by Holland et al. (1992) is a bio-inspired method widely used to solve complex optimization problems. GA has been shown to outperform classical non-evolutionary methods like Sequential Floating Search (Kudo & Sklansky, 2000), and Greedy-like Search (Vafaie, Imam, et al., 1994) to solve large-scale feature selection tasks. Moreover, better instances reduction rates and higher classification accuracy were obtained under experimental conditions using an instance selection GA in comparison with several experiments using non-evolutionary methods (Cano, Herrera, & Lozano, 2003).

High computational cost is however a major drawback of using GA for feature selection. Typically used as a wrapper method, the process of GA involves a large number of evaluations that are computationally heavy, particularly with on data sets containing a large number of instances. As a result, research employing GA for feature selection have been mostly limited to data sets with a small number of instances, i.e., less than 1000 (Xue, Zhang, Browne, & Yao, 2015).

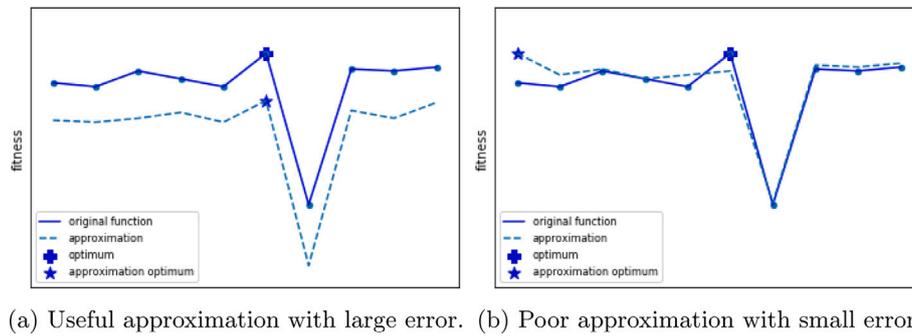

(a) Useful approximation with large error.  (b) Poor approximation with small error.

**Fig. 1.** A qualitatively useful approximation for a combinatorial optimization is shown in (a). The approximation correctly identifies the maximum of the original function, even though the approximation error is large. On the other hand, the approximation in (b) offers better quantitative approximation (values closer to the original fitness), but it leads to a false optimum.

In this work, we use a light-weight approximation of the computationally expensive GA fitness function to guide the feature selection task. We propose a novel active sampling method to construct high quality approximations. It is based on a concept introduced first by Jin, Hüsken, Sendhoff, et al. (2003), who used "qualitative" to define a useful approximation. This meta-model ranks different individuals similarly to the original fitness function, but not necessarily reproduces the exact value. An example of a useful qualitative approximation constructed using our method is shown in Fig. 1(a). The values of the approximate meta-model are consistently significantly lower than those of the original function. It preserves the qualitative properties of the original function in terms of relative fitness of different solutions. Therefore, the meta-model can be useful to lead the evolutionary optimization irrespective of the quantitative measures such as mean error. We define "Approximation Usefulness" and use the expected value of Spearman rank correlation ($\rho$) (Spearman, 1910, 1961) between the original function and meta-model evaluations as a quality measure of the meta-model.

An approximate classifier is trained with a subset of samples selected by our novel informed selection method. The accuracy of this surrogate model is used in the evolutionary computation of feature selection on data sets which are orders of magnitude larger than previous work reported in the literature. In our experiments, we tested the proposed method using data sets with hundreds of thousands of instances, and thousands of features. We have shown empirically that our algorithm $CHC_{QX}$ scales better than its classical wrapper counterpart, CHC, for larger datasets. Our algorithm is shown to converge faster to better feature subset solutions for datasets larger than 10K instances and was always significantly superior for datasets larger than 100K. We also have demonstrate the applicability of this approach to another class of EC by observing similar results with an algorithm based on PSO, namely $PSO_{QX}$.

## 2. Related work

The survey undertaken by Jin (2005) established two major concerns of approximating fitness evaluation in evolutionary computation. First, the approximation must ensure that the evolutionary algorithm convergences to a global optimum, or near-optimum, of the original fitness function. Second, the computational cost should be reduced. A subsequent survey (Jin, 2011) highlighted the limited success achieved in applications of meta-model based on evolutionary optimization even given what had been considerable growth in interest in using such meta-models within the research field.

Evolutionary feature selection methods can be grouped into three types: filter, hybrid and wrapper methods. Filters are computationally more efficient comparing to wrapper algorithms as they utilize measures such as correlation (Hall et al., 1999), mutual information (Jha & Saha, 2021; Zhou, Wang, & Zhu, 2022), ReliefF (Sun, Yin, Ding, Qian, & Xu, 2020), fisher score (Gu, Li, & Han, 2012), inconsistency rate (Lanzi, 1997; Liu, Setiono, et al., 1996) or even an ensemble of such measures (Ghosh et al., 2019) to estimate the fitness of feature subsets. Filter methods, however share the fundamental limitation of being agnostic towards the Machine Learning (ML) algorithm (Zhang, Li, Wang, & Zhang, 2013). Whereas wrappers evaluate the feature subsets based on the induction algorithm performance (Maldonado, Riff, & Neveu, 2022), which often results in better performance (Altarabichi, Fan, Pashami, Mashhadi and Nowaczyk, 2021; El Aboudi & Benhlima, 2016; Jović, Brkić, & Bogunović, 2015). In a review of 22 different filter methods (Bommert, Sun, Bischl, Rahnenführer, & Lang, 2020) concluded that no filter method outperforms all other methods on a consistent basis.

Hybrid methods combine a filter with a wrapper in a two-staged approach. In a hybrid a filter is applied first to the features with the goal of reducing the search space. Only the top ranked-features are used by the meta-heuristic in the second stage. This approach of filtering low-ranked features was used by Oreski and Oreski (2014), Rani, Kumar, Jain, and Chawla (2021), Song, Zhang, Gong, and Gao (2021), Sun, Jin, Xu, and Cichocki (2021) and Tan, Fu, Zhang, and Bourgeois (2008). Such approaches however suffer from two major drawbacks. First, the reduction in search space is only applicable to features. Therefore, a data set with large number of instances would not benefit much from this filter technique. Secondly, low-ranked features might turn out to be important when combined with other features. This filter-based approach also misses any potential feature interactions. Some hybrid approaches overcome this by employing local search along with the binary optimization algorithms (Chattopadhyay, Kundu, Singh, Mirjalili, & Sarkar, 2022; Ghosh, Malakar, Bhowmik, Sarkar, & Nasipuri, 2017; Kabir, Shahjahan, & Murase, 2011). It remains challenging however to determine how much such approaches are compromising the accuracy to reduce computation time without a comparison against a wrapper. As such methods are often only compared with baseline models and other hybrid and filter methods.

The final approach to feature selection is the wrapper, and our method belongs to this category. Wrapper approaches rely on the ML model to explicitly evaluate the fitness of feature subsets. The first work to address the computational cost of GA for the feature selection task using an approximate model in a wrapper settings was Brill, Brown, and Martin (1992). They proposed two key ideas: first, a simple k-Nearest Neighbours (kNN) is used to approximate fitness evaluations of a Neural Network. Secondly, they proposed a method named "training set sampling", in which only a portion of the training set is used to train a model for GA evaluations. However, their method was only validated using one small data set of 30 features and 150 of instances. Also, re-sampling on each generation using this training set sampling method forced more evaluations, and consequently incurred a high computational cost. A recent work by Altarabichi, Nowaczyk, Pashami and Mashhadi (2021) combined the "training set sampling" concept with ideas from progressive sampling to create a multi-level surrogate-assisted algorithm that outperformed a wrapper GA by both being faster to converge and by leading to feature subset solutions of higher accuracy.

Progressive Sampling used by Le, Van Tran, Nguyen, and Nguyen (2015) to identify the Optimal Sample Size (OSS) which has been defined as the smallest sample size that offers minimum achievable error for a given learning algorithm. A model trained with OSS was used in fitness function evaluations of GA. Additionally, Le et al. proposed parallelization of the fitness function computation to reduce the runtime of GA. The algorithm was used to perform feature selection for the Named Entity Recognition (NER) task.

A Coevolutionary approach to performing feature and instance selection simultaneously for data set reduction showed promising results over a wide range of data sets. The largest data set used, in terms of number of instances, however, had 1728 instances. Whereas the largest in terms of features, had only 60 features (Derrac, García, & Herrera, 2009).

A GA feature selection method based on the MapReduce paradigm was proposed by Peralta et al. (2015). The algorithm in this method decomposed the original data set into blocks of instances to learn from in the map phase; then, the obtained partial results are merged into a final vector of feature weights in the reduce phase. The selected features are identified using a threshold applied to feature weights vector. Although the work was sufficiently convincing to show the usefulness of the MapReduce paradigm in reducing the computational cost, the performance of the method was only compared against baseline models trained with all features for two large data sets.

## 3. Problem formulation

In this section we further formally define the feature selection problem. We also offer a fundamental overview of using GA as a feature selection method in a wrapper setting, and its computational challenges. In the second part of this section, we discuss approximating the fitness function using a quantitative approximation and demonstrate the major obstacles of constructing a light-weight approximation following a quantitative approach.

### 3.1. Feature selection using genetic algorithm

We start by defining the feature selection problem for a machine learning task. We are given a data set D of labelled pairs of the dimensions $(n \times k)$, of which $n$ represents the number of instances and $k$ is the number of features. An instance $\vec{x}$ can be expressed as a $k$-dimensional real-valued vector $\vec{x} \in \mathbb{R}^k$. The goal of the feature selection task is to select a new subspace $\mathbb{R}^l$ from $\mathbb{R}^k$ (where $l \leq k$), while maintaining a comparable (or even better) performance to the one obtained with the original feature space $\mathbb{R}^k$. An instance $\vec{x'}$ after feature selection can be expressed as a $l$-dimensional real vector $\vec{x'} \in \mathbb{R}^l$.

Typically, a GA used for feature selection is initiated with a random population of individuals encoding feature subsets as chromosomes of binary strings. An individual $g$ can be expressed as $g \in \{0, 1\}^k$, where 1 indicates the selection of feature of the corresponding index, while 0 indicates exclusion. As a wrapper method, GA evaluates individual's fitness by constructing a classification (or regression) model using the feature subset represented by this individual's chromosome.

The algorithm proceeds from one generation to the next by applying crossover and mutation operators to the selected individuals to produce offspring. The process resembles natural selection in that it chooses individuals with the highest level of fitness as the most likely to propagate to the following generation. In this work, we have used the CHC Genetic Algorithm (Whitley & Sutton, 2012) to lead our search process as it has been shown to perform well with small populations (Eshelman, 1991), whilst being more computationally efficient. Whenever a GA is mentioned in the rest of this paper, it is always CHC-based.

Finding the optimal features subset of high-dimensional data sets, even for a small population, requires running a large number of fitness function evaluations (for the experiments in Section 5.2 it varied from 1002 to 2216) — where each evaluation is computationally expensive. The computational cost of GA is linearly dependent on the time complexity of the model used to evaluate fitness of different feature subsets (Altarabichi, Nowaczyk et al., 2021). The time complexity of many classification algorithms as a function of the number of samples is $O(n^c)$, where $(c \geq 1)$. For example the training complexity of kNN is $O(n^2 k)$ (Brill et al., 1992), while nonlinear SVM is between $O(n^2)$ and $O(n^3)$ (Bottou & Lin, 2007). Consequently, the complexity order of a wrapper feature selection GA using kNN is $O(n^2 k t)$, and is between $O(n^2 t)$ and $O(n^3 t)$ for SVM, where t is the total number of fitness function evaluations.

As the feature selection *GA* is quadratic or even cubic with respect to the number of training instances, several methods utilized processing models trained with smaller samples for fitness evaluations. We categorize these methods under three main categories: training set sampling (Altarabichi, Nowaczyk et al., 2021; Brill et al., 1992; Le et al., 2015), dividing the dataset into small pieces (Peralta et al., 2015) and instance selection (Derrac et al., 2009). Our method belong to the last category as we speed up the computation by reducing the number of training instances. Several other methods suggested reconstructing the training set through instance selection. Liu, Wang, Wang, Lv, and Konan (2017) identified useless instances that are unlikely to be support vectors to make the training time of SVM manageable. Saha, Sarker, Al Saud, Shatabda, and Newton (2022) selected instances in unsupervised fashion from the centre and borders of the clusters found using K-Means algorithm. Shaw et al. (2021) proposed an instance selection algorithm to select instances in class imbalance settings.

Our approach can be distinguished from all mentioned methods that rely on quantitative measures (e.g., accuracy of the model trained with selected instances) to evaluate the goodness of the resulting training set. We select instances that lead to correct selection of solutions during feature selection.

### 3.2. The drawbacks of a quantitative approximation approach

In this section we highlight the major drawbacks of constructing an approximation of the original function following a quantitative approach. A quantitative approximation offers a small approximation error when compared to the original function. In the context of approximating the fitness function of the feature selection task, a natural choice would be classifier $C^{oss}$ trained with the optimal sample size OSS, as $C^{oss}$ offers a close quantitative approximation of $C^D$ trained with the complete data set D of n instances.

The first possible drawback of following a quantitative approach is evident in Fig. 1(b), in which a meta-model with small approximation error fails to guide evolutionary computations properly and leads to a false optimum. A false optimum is defined as a point in the optimization surface that corresponds to an optimum[3] of the approximate function, but not of the original fitness function (Jin, 2005). Convergence to such false optima in the approximate model is a major problem in surrogate-assisted evolutionary optimization (Jin, Olhofer, & Sendhoff, 2000). The severity of the problem depends, of course, on how close the real (original fitness) optimum is to the false optimum — however, this is impossible to know at the start, and even for an approximation with small degree of error this can be arbitrarily distant.

---

[3] A false optimum could either correspond to a maximum or a minimum, based on the nature of the optimization task.

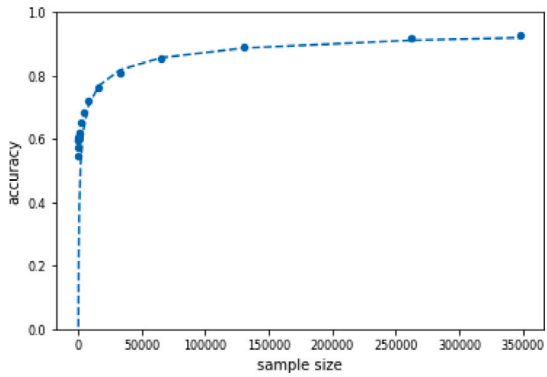

**Fig. 2.** The Learning Curve of the `covtype` data set using a Decision Tree model, and a Geometric Sampling Schedule $S_g = \{32, 64, 128, \ldots, 262\,144\}$. The last point to the right represents training with all the available 348 861 training set instances.

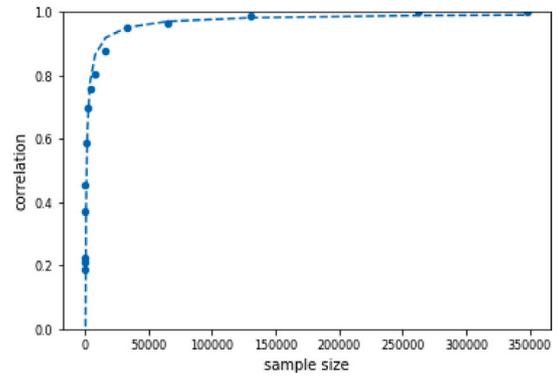

**Fig. 3.** The "Approximation Usefulness" curve of the `covtype` data set using a Decision Tree model, and a Geometric Sampling Schedule $S_g = \{32, 64, 128, \ldots, 262\,144\}$. The points correspond to the rank correlation scores between the original classifier and an approximation trained using the corresponding sample size sampled randomly.

Additionally, there are no guarantees that the OSS will correspond to a small number of instances. The performance of most ML models, especially Deep Learning (DL) but also many simpler ML models, does not converge quickly, and continues to improve significantly with more labelled data. Progressive sampling (PS) could be used to establish the relation between sample size and model accuracy using batches that incrementally grow in size. The relation can be depicted with a learning curve showing model accuracy as a function of sample size. A well-behaved learning curve usually follows an inverse power law function (Yelle, 1979). Fig. 2 shows the learning curve of the covtype data set from the UCI ML repository. The sample batch in the figure increases in size following a Geometric Schedule (Provost, Jensen, & Oates, 1999) by the equation $S_g = a^i \cdot n_0$. We may observe from Fig. 2 that the convergence point is not reached, for a simple Decision Tree classifier, even when training on more than 250$K$ instances. Therefore, our goal is to generally identify a much smaller sample size (compared to OSS) to train a meta-model used for the feature selection task.

## 4. Method

Our method section is divided into two subsections. In the first one, we formulate the definition of "Approximation Usefulness" and introduce the concept of measuring the value of qualitative approximation. In the subsequent subsection, we present our method of actively sampling instances with the purpose of creating a light-weight meta-model that, to a high degree, satisfies the "Approximation Usefulness" definition.

### 4.1. Approximation usefulness

In our work, we construct a classifier $C^s$ trained with a particular subset of samples ($s$), where $s \subset D$ and $|s| \ll n$, to carry out the FS task. Our approach exploits the idea highlighted by Jin et al. (2003), that from an evolutionary computation perspective, the quantitative quality of the approximation is irrelevant, only the correct selection must be ensured. In our FS problem, the approximate classifier $C^s$ must rank different feature solutions similar to $C^D$ to be considered a useful approximation. However, its overall accuracy can be much lower. We therefore define "Approximation Usefulness" to indicate the necessary conditions that must be satisfied to ensure correctness of the GA computations when an approximation is used.

**Definition 1.** A meta-model is sufficient to lead the evolutionary computations to the correct maximum[4] of the fitness function if the following conditions are satisfied:

$$\forall g_1, g_2 : f^D(g_1) > f^D(g_2) \implies f^s(g_1) > f^s(g_2) \quad (1)$$

$$\forall g_1, g_2 : f^D(g_1) = f^D(g_2) \implies f^s(g_1) = f^s(g_2), \quad (2)$$

where $g_1$ and $g_2$ are two possible solutions (individuals), $f^D(g)$ is the fitness value of individual ($g$) using the original fitness function (a classifier trained with the complete data set $D$ of $n$ instances), $f^s(g)$ is the fitness value of individual $g$ using the approximated fitness function (a classifier trained with the subset $s$ of instances in $D$).

It is important to highlight that Eqs. (1) and (2) are sufficient conditions to realize a useful approximation, irrespective to how good the approximate classifier $C^s$ is in performing the learning task. In other words, $C^s$ could have a low accuracy[5] in comparison to $C^D$, but it can still be considered a useful approximation of $C^n$ in the feature selection task, provided both classifiers rank different feature subsets similarly. The Spearman Rank correlation was used in Jin et al. (2003) as a metric to measure the quality of meta-models. We will similarly use rank correlation as a quantitative measure of how valid is the qualitative approximation in satisfying the Approximation Usefulness conditions.

To visualize the qualitative aspect of the problem, we plot in Fig. 3 what we define as the Approximation Usefulness Curve. Contrary to the Learning Curve we have observed in Fig. 2, the $y$-axis for the Approximation Usefulness plot represents the expected value of rank correlation between evaluations done using the original function (trained with all data) and the same evaluations done using approximations (trained with progressively larger sample size). Intuitively, we would expect the learning algorithm to exploit more complex interactions between features as more data become available.

A meta-model trained with a very small sample of instances could rank different feature subset solutions differently than the same learning algorithm trained with all the data. As we allow the meta-model to progressively access larger samples, we would expect the meta-model to have better agreement with the original function on the rank of feature subsets.

The goal of the Approximation Usefulness Curve is, therefore to capture the relation between training the meta-model with progressively larger samples and the agreement with original function evaluations. We estimate the expected value of rank by randomly generating $q$ feature subsets and evaluating these subsets using the original function, and meta-models with progressively larger samples.

---

[4] The definition applies as well in the case of a minimization problem. In this work however, we are always working with finding maximum.

[5] The concept generalizes directly to any other metric of measuring classifier performance (e.g., recall or AUC). In this work, for simplicity, all evaluations are based on classifier accuracy.

The vector $\vec{u} = (u_1, u_2, \ldots, u_q)$ represents the $q$ randomly generated feature subsets, where $u_i \in \{0,1\}^k$. Consequently, for any sample size $m$, we create two $q$-dimensional vectors $\vec{o} = (o_1, o_2, \ldots, o_q)$, and $\vec{a}^m = (a_1^m, a_2^m, \ldots, a_q^m)$, where $o_i$ is the fitness value of $u_i$ calculated using the original fitness function, and $a_i^m$ is the fitness value of $u_i$ calculated using an approximation trained with a sample of size $m$ randomly sampled from D. Accordingly, we depict approximation usefulness as a function of sample size using progressive sampling. We calculate the rank correlation of the two vectors $\rho(\vec{o}, \vec{a}^m)$ using values of $m$ that follow a geometric sampling schedule.

We may observe from Fig. 3 that already an approximation trained with a random sample of size 32 768 (less than 10% of available training instances) of the covtyp data set exceeds 0.95 correlation. Such meta-model will mostly make correct selections, despite the large difference in accuracy between the original and approximate models as observed from Fig. 2; a model trained with all 348 861 training examples achieved 93.00% accuracy on test set, in comparison to 80.60% for a model trained with 32 768 samples.

### 4.2. The $CHC_{QX}$ algorithm

Our approach is based on the hypothesis that well-informed instance selection can produce high quality meta-models using smaller sample sizes and it is enough to observe few evaluations of the original function to identify such key instances. Active selection of samples is shown to improve model quality significantly in different research areas (Jin, 2005). We propose $CHC_{QX}$, **CHC Q**ualitative appro**X**imation, a two-staged surrogate-assisted evolutionary algorithm for feature selection. In our algorithm (the full pseudocode is in Algorithm 3), we break the optimization problem of feature selection into two parts. In the first part, we use active selection of samples to construct high quality lightweight meta-model. Once a good meta-model is constructed, we use it to solve the second optimization problem, i.e., finding the best possible feature subset.

#### 4.2.1. $CHC_{QX}$ active sampling phase

Formally, this optimization problem aims to find a small subset of instances $s$ from data set D (where $|s| \ll n$), which maximizes the expected correlation between $C^s$ and $C^D$. A meta-model trained with $s$ is expected to rank feature subsets similarly to the original function. The data set $D'$ after instance selection can be expressed as matrix $(|s| \times k)$. $CHC_{QX}$ uses an instance selection GA to solve this optimization problem. The instance selection GA is initialized with a population of individuals $g \in \{0,1\}^n$, where 1 indicates the selection of the instance of the corresponding index.

To measure the quality of a candidate meta-model during the instance selection phase, we first randomly generate $q$ feature subsets. The randomly generated solutions serve as snapshots of the optimization surface of the original function. Intuitively, $CHC_{QX}$ tries to construct a meta-model with a fitness landscape that is both aligned, and highly correlated with, the optimization surface of the original function as in Fig. 1(a).

As we did in 4.1, we denote $\vec{u}$ as the vector of randomly generated feature subsets, $\vec{u} = (u_1, u_2, \ldots, u_q)$, where $u_j \in \{0,1\}^k$, $j \in \{1, 2, \ldots, q\}$. We calculate the q-dimensional vector $\vec{o} = (o_1, o_2, \ldots, o_q)$, where $o_i$ is the fitness value of the corresponding feature subset calculated using the original fitness function. Accordingly, $CHC_{QX}$ instance selection tries to find the instance subset that maximize the expected rank, while minimizing the number of selected instances according to the following fitness function:

$$f_{is}(g) = (1 - \rho(\vec{O}, \vec{A})) + \frac{|s|}{n} = \frac{6\sum_i d_i^2}{q(q^2-1)} + \frac{|s|}{n} \quad (3)$$

where $\vec{a} = (a_1, a_2, \ldots, a_q)$ is a $q$-dimensional vector of fitness value of the corresponding randomly generated feature subset calculated using a classifier trained with instances identified by individual $g$, and $d_i$ is the difference in rank of $r_i$ (the $i$th feature subset) between $\vec{o}$ and $\vec{a}$. Values of $\rho(\vec{o}, \vec{a})$ fall in the range +1 to −1, where the maximum value of +1 indicates an optimal approximation, with Eqs. (1) and (2) satisfied.

By minimizing the fitness function defined in (3), the instance selection of $CHC_{QX}$ optimizes for higher quality meta-model using the left term of the fitness function, with the smallest number of instances (as captured by the right term). The fitness of the original function (an individual trained with all $n$ available instances in D), according to (3), is equal to $f_{is}(g) = (1-1) + \frac{n}{n} = 1$. A high-quality candidate approximation trained with small number of instances will have a fitness below 1 and approaches 0.

We denote $s^*$ as the instance subset that results from solving the first optimization problem. We use $s^*$ to train the approximate classifier $C^*$ that will be used next in the feature selection phase.

#### 4.2.2. $CHC_{QX}$ feature selection phase

During the feature selection phase, our algorithm uses the approximate classifier $C^*$ that was constructed in the instance selection phase, together with the original fitness function $C^D$. We carry majority of feature subset evaluations using the approximation $C^*$, and only after a fixed number of generations we reevaluate all individuals using $C^D$. We control this frequency of using the original function through the frequency hyper-parameter ($f$). The use of the approximate model together with the original fitness function is known as evolution control in evolutionary computations using approximation (Jin et al., 2000), and has been recognized for its effectiveness in preventing the approximation from converging to a false optimum (Ratle, 1998).

The fitness function of the second optimization problem is given by the equation:

$$f_{fs}(g) = Acc_{val}(g) \quad (4)$$

where $Acc_{val}$ is the validation set accuracy of classifier $C^*$ when trained with feature subset $g$. We denote $g^*$ as the feature subset that results from solving the second optimization problem. The feature subset $g^*$ is the final solution of $CHC_{QX}$.

The pseudocode of the instance selection and feature selection stages of $CHC_{QX}$ can be found in Algorithm 1 and Algorithm 2. We explain the fundamental steps of the CHC algorithm as it is used in both stages:

1. **Initialization**: The initial population $P_0$ is generated randomly according to the hyper-parameters $e$ that identifies the number of individuals in the population, and $pr$ that indicates the independent probability of having a 1 in each bit of the string.
2. **Reproduction selection**: On each generation t, the parents $C_t$ are selected randomly for reproduction from the population $P_t$. But an incest prevention mechanism is applied to prevent similar parents from mating. Similarity is identified by measuring the hamming distance between the pair of parents and only pairs which differ from each other by the threshold given by $d$ are allowed to mate.
3. **Heterogeneous recombination**: The HUX operator is used to generate off-springs $C'_t$ by coping all bits matched in both parents, and then copying half of the different bits from each parent to the resulting off-springs.
4. **Cross generation elitist selection**: The algorithm selects the best individuals from the current generation $P_t$ and the off-springs $C'_t$ to be passed to the following generation. This cross generation elitist strategy ensures that the best solutions found so far always survive. If the stop criterion is met after this step, the algorithm simply exits and returns the best found solution. Alternatively, the process moves to step 5 if the progress stagnate for several generations. Otherwise, the algorithm goes back to steps 2 and steps 2–4 are repeated iteratively.

5. **Cataclysmic mutation (restarts)**: Mutation is not used in the recombination stage of CHC. It is only used whenever convergence is reached based on the $d$ value. A restart is initiated to reintroduce diversity into the population when the progress stagnate for several generations. The best individual is used as a template to generate the new population by mutating 35% of its bits. The process goes back to step 2 after the restart.

Refer to Algorithm 3 for the full pseudocode of $CHC_{QX}$ feature selection. The main steps of Algorithm 3 are:

1. **Instance selection**: This step represents the first stage of the $CHC_{QX}$ algorithm and is carried to identify the meta-model $C^*$ training instances.
2. **Feature selection using meta-model**: Feature selection is carried using $C^*$ for a fixed number of generations according to the $f$ value.
3. **Reevaluations using original function**: Every $f$ generations the whole individuals in the feature selection population $P_t$ are reevaluated using the original function $C$.
4. **Stop criterion**: Once the stop criterion is met, the algorithm returns the best found feature subsets $g^*$. Otherwise, we go back to step 2.

## 5. Results and discussion

This section describes the experimental design we have used to evaluate our method, along with the results obtained in several experiments. The first experiment aimed to evaluate $CHC_{QX}$ against a traditional CHC, an algorithm that uses all the available training data for feature selection. We also report the results of another proposed algorithm, $PSO_{QX}$, and compare it to PSO. In the second and third experiments we conducted sensitivity analysis of the main hyperparameters of the algorithm, namely we evaluated the effect of varying the population size and the frequency of the evolution control hyperparameters on $CHC_{QX}$ performance. Finally, we provide amortized analysis of the complexity time of the $CHC_{QX}$ algorithm and compare it to CHC.

### 5.1. Experimental setup

Our experiments involve 13 data sets from the UCI Machine Learning Database Repository. We have included 6 small sized datasets of less than 10K, 4 medium size datasets between 10K and 100K, and 3 large datasets with more than 100K instances. The objective was to evaluate the effectiveness of our approach for datasets of varying sizes. Table 1 provides a summary of the number of instances, features and classes of all data sets used in our experiments.

Each data set is divided into training (60%), validation (20%), and testing (20%) splits. We have used a Decision Tree classifier in all experiments with the default algorithm settings of the Python library (sklearn[6]) implementation. A unified approach of prepossessing is adopted for all data sets, including categorical features encoding, imputation of missing values, and shuffling of instances. Accuracy of the model is the metric we used for the evaluation in all experiments. All reported accuracies are the ones realized on the testing set. The

---

[6] https://scikit-learn.org/stable/.

**Algorithm 1:** $CHC_{QX}$ Active Sampling

**Input** : ($q$) number of controlled individuals.
($e$) number of individuals in population.
($t_{max}$) maximum number of generations.
($pr$) probability of each instance to be selected (independently).
($d$) diversity parameter.
($div$) divergence rate on restarts.

**Output:** ($s^*$) instances selected to train the meta-model.

// Create vector $u$ of $q$ randomly generated feature subset.
1  $Generate[u, q]$;
   // Evaluate the $q$ feature subsets within u using the original fitness function.
2  $Evaluate[u]$;
   // Initialize the generation counter $t$ to zero.
3  $t \leftarrow 0$;
   // Initialize a population of $e$ instance individuals using $pr$.
4  $Initialize[P_t, e, pr]$;
   // Evaluate the initial population.
5  $Evaluate[P_t]$;
6  **while** $t < t_{max}$ **do**
       // Select parents randomly.
7      $C_t \leftarrow Select[P_t]$;
       // Mate parents with incest prevention.
8      $C'_t \leftarrow HUX[C_t, d]$;
       // Evaluate offsprings.
9      $Evaluate[C'_t]$;
       // Pass best $e$ individuals from parents and offsprings to the following generation.
10     $P_t \leftarrow Select[P_t, C'_t, e]$;
       // Increase the generation counter by one.
11     $t \leftarrow t + 1$;
12     **if** $Convergence[P_t]$ **then**
           // Population is reinitialized as CHC converged.
13         $P_t \leftarrow Restart[P_t, div]$
   // Select the best instance subset.
14 $s^* \leftarrow Select[P_t, 1]$;
15 **return** $s^*$

hardware used for all experiments is a laptop with 1.6 GHz processor and 8 GB of memory.

In describing the experiments, we will use the following terminology to refer to different classifiers:

- **Baseline DT:** The Decision Tree classifier that is trained on all available instances without performing any feature selection.
- **CHC:** The Decision Tree classifier that is trained on all available instances in the training set, after performing feature selection using CHC with the original fitness function. The population size of CHC is 50 in all experiments unless explicitly stated otherwise, while other parameters are set to the recommended settings suggested in the original paper (Eshelman, 1991). The diversity parameter is set to ($d = \frac{k}{4}$), where $k$ is the length of the individual (number of features), while the divergence rate is ($div = 0.35$).
- **PSO:** The Decision Tree classifier that is trained on all available instances in the training set, after performing feature selection using PSO with the original fitness function. We have used the global version of PSO with a topology connecting all particles to one another. The following options are used {c1: 1.49618, c2: 1.49618, w: 0.7298}, while the number of particles is set to 50 in all experiments.

**Algorithm 2:** CHC Feature Selection

   **Input** : (*model*) model to be used for evolutionary evaluations.
           (*e*) number of individuals in population.
           ($t_{max}$) maximum number of generations.
           (*pr*) probability of each feature to be selected (independently).
           (*d*) diversity parameter.
           (*div*) divergence rate on restarts.
   **Output:** $g^*$ Best found feature subset.
   // Initialize the generation counter *t* to zero.
1   $t \leftarrow 0$;
   // Initialize a population of *e* instance individuals using $pr < 0.5$.
2   $Initialize[P_t, e, pr]$;
   // Evaluate the initial population.
3   $Evaluate[P_t, model]$;
4   **while** $t < t_{max}$ **do**
      // Select parents randomly.
5     $C_t \leftarrow Select[P_t]$;
      // Mate parents with incest prevention.
6     $C'_t \leftarrow HUX[C_t, d]$;
      // Evaluate offsprings.
7     $Evaluate[C'_t]$;
      // Pass best *e* individuals from parents and offsprings to the following generation.
8     $P_t \leftarrow Select[P_t, C'_t, e]$;
      // Increase the generation counter by one.
9     $t \leftarrow t + 1$;
10    **if** $Convergence[P_t]$ **then**
        // Population is reinitialized as CHC converged.
11       $P_t \leftarrow Restart[P_t, div]$
   // Select the best feature subset.
12   $g^* \leftarrow Select[P_t, 1]$;
13   **return** $g^*, P_t$

**Algorithm 3:** $CHC_{QX}$ Feature Selection

   **Input** : (*q*) number of controlled individuals.
           ($e_1$) number of individuals in instance selection population.
           ($e_2$) number of individuals in feature selection population.
           ($t_{max1}$) maximum number of generations during instance selection.
           ($t_{max2}$) maximum number of generations during feature selection.
           ($pr_1$) probability of each instance to be selected (independently).
           ($pr_2$) probability of each feature to be selected (independently).
           (*f*) frequency of using the original function.
           (*d*) diversity parameter.
           (*div*) divergence rate on restarts.
   **Output:** $g^*$ best found feature subset.
   // Select meta-model training instances from available data.
1   $s^* \leftarrow CHC_{QX}ActiveSampling[q, e_1, t_{max1}, pr_1, d, res]$;
   // Initialize the feature selection generation counter *t*.
2   $t \leftarrow 1$;
3   **while** $t < t_{max2}$ **do**
4     **if** $t \bmod f \neq 0$ **then**
        // Carry feature selection for one generation using $C^*$.
5       $g^*, P_t \leftarrow CHCFeatureSelection[C^*, e_2, 1, pr_2, d, div]$;
6     **else**
        // Reevaluate the population using the original function *C*.
7       $Evaluate[P_t, C]$;
      // Select the best feature subset.
8       $g^* \leftarrow Select[P_t, 1]$;
      // Increase the generation counter by one.
9     $t \leftarrow t + 1$;
10   **return** $g^*$

**Table 1**
UCI data sets used for evaluation.

| Data set | No. of instances | No. of features | No. of classes |
| --- | --- | --- | --- |
| dermatology | 366 | 34 | 6 |
| german | 1000 | 24 | 2 |
| semeion | 1592 | 265 | 2 |
| car | 1728 | 6 | 4 |
| abalone | 4177 | 8 | 28 |
| qsar | 8992 | 1024 | 2 |
| adult | 32 561 | 14 | 2 |
| bank-full | 45 211 | 16 | 2 |
| connect-4 | 67 556 | 42 | 3 |
| dota2Train | 92 650 | 116 | 2 |
| diabetes | 101 766 | 49 | 3 |
| census-income | 199 523 | 41 | 2 |
| covtype | 581 012 | 54 | 7 |

- **$CHC_{QX}$**: The Decision Tree classifier that is trained on all available instances in training set after performing feature selection using $CHC_{QX}$. This algorithm uses the same settings of the base optimizer as the CHC baseline. The hyper-parameters specific to $CHC_{QX}$ are set to $q = 20$ and $f = 10$.
- **$PSO_{QX}$**: The Decision Tree classifier that is trained on all available instances in the training set after performing feature selection using $PSO_{QX}$. This algorithm use the same settings of the base optimizer as the PSO baseline. The hyper-parameters specific to $PSO_{QX}$ are also set to $q = 20$, and $f = 10$.

### 5.2. Experiment I: $CHC_{QX}$ vs. CHC and $PSO_{QX}$ vs. PSO

In this experiment, we compared our algorithm against a classical wrapper feature selection method. The objective was to demonstrate the effectiveness of $CHC_{QX}$ using datasets of varying sizes in terms of number of instances and features. To validate whether $CHC_{QX}$ converges faster than CHC, we ran $CHC_{QX}$ to convergence (defined as 10 consecutive generations with no improvement for best solution in population) and allowed CHC to run for the same CPU time. Due to the stochastic nature of the process, we ran 10 repetitions of each dataset and reported the median and standard deviation of the runs.

The results in Table 2 show that as expected, in comparison to a baseline without any feature selection, all feature selection methods improved the performance significantly, based on paired Student's *t*-test. It must be noted however, that our algorithms $CHC_{QX}$ and $PSO_{QX}$ were so successful in reducing overfitting of the baseline Decision Tree to the degree that the DT model after feature selection managed to exceed the performance of a Random Forest model trained with all features (ensemble of 11 Decision Trees) for a number of datasets (abalone, adult, dota2Train, diabetic and covtype).

The results of Table 2 show no advantage of using $CHC_{QX}$ for small datasets with less than 10K instances. This is quite natural and can be explained by the overhead of the algorithm, which caused a deteriorating effect in a number of datasets (semeion, abalone and qsar).

**Table 2**
The median and standard deviation results of 10 runs, bold indicates significance in comparisons between $CHC_{QX}$, $PSO_{QX}$ against CHC and PSO, respectively.

| Dataset | Accuracy | | | | |
|---|---|---|---|---|---|
| | Baseline DT | CHC | $CHC_{QX}$ | PSO | $PSO_{QX}$ |
| dermatology | 95.96% | 93.24% ± 1.94 | 93.24% ± 3.04 | 95.95% ± 1.70 | 95.95% ± 1.72 |
| german | 66.00% | 68.00% ± 3.25 | 70.00% ± 3.52 | 66.75% ± 3.10 | **71.04%** ± 4.21 |
| semeion | 93.42% | 93.73% ± 1.20 | 93.26% ± 1.61 | 93.42% ± 1.11 | 92.63% ± 0.78 |
| car | 98.27% | 98.27% ± 0.00 | 98.27% ± 0.00 | 98.27% ± 0.00 | 98.27% ± 0.00 |
| abalone | 19.98% | 25.60% ± 2.35 | 24.82% ± 3.09 | 22.07% ± 2.82 | 19.86% ± 3.51 |
| qsar | 89.05% | 90.94% ± 0.48 | 90.69% ± 0.54 | 90.86% ± 0.68 | 91.19% ± 0.82 |
| adult | 80.58% | 85.12% ± 0.04 | 85.12% ± 0.00 | 85.11% ± 0.35 | 85.19% ± 0.27 |
| bank-full | 86.77% | 89.16% ± 0.06 | **89.22%** ± 0.06 | 89.26% ± 1.74 | 89.30% ± 0.07 |
| connect-4 | 72.00% | 75.92% ± 1.26 | 76.68% ± 0.76 | 75.60% ± .90 | **76.54%** ± 0.67 |
| dota2Train | 51.78% | 53.85% ± 0.74 | 54.04% ± 0.87 | 53.89% ± 0.48 | **54.79%** ± 0.83 |
| diabetic | 49.21% | 55.97% ± 1.02 | **56.93%** ± 0.27 | 56.36% ± 0.68 | **57.04%** ± 0.37 |
| census-income | 92.87% | 94.63% ± 0.35 | **94.94%** ± 0.07 | 94.79% ± 0.24 | **94.94%** ± 0.06 |
| covtype | 92.99% | 93.36% ± 0.16 | **93.77%** ± 0.12 | 93.51% ± 0.15 | **93.73%** ± 0.08 |

However, the advantage of using our algorithm is already noticeable for medium size datasets (between 10K and 100K), as $CHC_{QX}$ performs generally better than CHC, and is significantly better according to the paired *t*-test for the `bank-full` dataset. This trend continues, and $CHC_{QX}$ is always significantly better than CHC for all three large datasets (`diabetic`, `census-income` and `covtype`), i.e., those with more than 100K instances. This demonstrates the usefulness of our approach for the purpose it was designed for, namely as the datasets scale larger in size.

We show similar results with our proposed PSO based algorithm. As we compare $PSO_{QX}$ against PSO following the same procedure. The results of Table 1 show $PSO_{QX}$ is generally no better than PSO for small sized datasets, while being generally better for medium sized, and is always significantly better for large datasets. This demonstrates that our procedure of constructing high quality meta-models is useful when used independently from the underlying optimizer.

*5.3. Experiment II: Effect of population size on $CHC_{QX}$ and CHC*

The objective of this experiment was to analyse the effect of increasing the population size for both CHC and our proposed algorithm $CHC_{QX}$. We vary the population size hyper-parameter to values equal to (50%, 100%, 200%, 400%) of the individual length and measure the impact on convergence time and fitness for both algorithms in 10 repetitions. Unlike the previous experiment, this time we let both algorithms run to convergence (defined as 10 consecutive generations with no improvement for the best individual). We studied the effect using two medium size datasets `adult` and `bank-full`, and one large dataset `census-income`.

An expected, we observe in Fig. 4 that the convergence time of both algorithms increases as we run the optimization with larger populations. It is clear, however, that $CHC_{QX}$ converges faster than CHC in almost all settings. In general, the fitness of both algorithms improves as we increase the population size. Also, agreeing with the results of Xu and Gao (1997), we observed a reduced possibility of premature convergence with larger populations. This was particularly apparent for CHC (as indicated by lower variations in the quality of final solutions). However, increasing the population size beyond 200% of the individual size does not seem to offer any significant improvements for either of the algorithms. This observation is consistent with results from the literature which suggests that large population size is not always helpful (Chen, Tang, Chen, & Yao, 2012) while other studies recommend using CHC with small populations of around 50 individuals (Whitley, 1994). We recommend, accordingly, to work with population size values between 100% and 200% of the individual size for both CHC and $CHC_{QX}$. A fixed value of 50 individuals performed consistently well for problems of varying complexity across our experiments and was therefore chosen as the default population size setting for $CHC_{QX}$.

*5.4. Experiment III: Effect of evolution control frequency on $CHC_{QX}$*

In this experiment, we analyse our evolution control strategy by varying the value of the frequency hyper-parameter. Intuitively, a trade-off exists between less frequent controls (which risk allowing the meta-model to mislead the optimization to a false optimum), or more frequent controls (which require increased computation time). We studied this trade off by varying the value of the *f* hyper-parameter between (5, 10, 20, 40). The value of *f* represents the number of generations that must pass before the original function is used during feature selection; a high value indicates less frequent use. We use the same three datasets as in the previous experiment.

We show in Fig. 5 the downside of less frequent use of evolution control for the value of (f = 40) in the `adult` dataset and (f = 20, f = 40) for the `bank-full` dataset. In these settings, the meta-model often converges to false optima, as is evident from the significant performance drop in comparison to the more controlled settings (f = 5, f = 10). Interestingly, the less controlled settings perform generally better for the `census-income` dataset; this observation could be explained by the fidelity of the approximate model. Intuitively, a high-quality meta-model requires less frequent evolution control (Jin, 2005). We recommend setting the value of *f* to be between 5 and 10, as this range provided the most well-balanced performance during our experiments. Ideally, a solution would be to abandon the fixed control strategy and follow an approach that adjust the frequency of evolution control adaptively, based on the fidelity of the meta-model (Jin, Olhofer, & Sendhoff, 2001), however this is work for the future and outside of the scope of this paper.

*5.5. Amortized analysis of the $CHC_{QX}$ algorithm*

As reduction of computation time is the main objective of our algorithm, we carried out amortized analysis of the time complexity of $CHC_{QX}$ in comparison to CHC. We showed that the amortized cost of one generation of $CHC_{QX}$ using the algorithm default settings is in the worst case smaller than its counterpart CHC, when the two algorithms run for more than 13 generations. We used the aggregate method to determine the upper bound of the worst case total run-time cost of *r* generations of evolution, then calculated the amortized cost of one generation for each method.

In our analysis both algorithms use a Decision Tree classifier with complexity $O(n \cdot k^2)$ (Su & Zhang, 2006), where *n* represents the number of instances and *k* is the number of features. The analysis could be directly extended to any other induction algorithm with a different complexity.

The time complexity of CHC is linearly dependent on the complexity of the induction algorithm. This is true for situations in which fitness evaluations consumes almost all the run-time of the algorithm, the time complexity of running the evolutionary operators like crossover and mutation is negligible in comparison. The time complexity of CHC could be expressed for Decision Tree as $O(n \cdot k^2 \cdot e \cdot r)$, where $e$ is the number of fitness function evaluations per generation.[7] The amortized cost of one generation of evolution using CHC is simply:

$$\frac{T_{CHC}(r)}{r} = \frac{n \cdot k^2 \cdot e \cdot r}{r} = n \cdot k^2 \cdot e$$

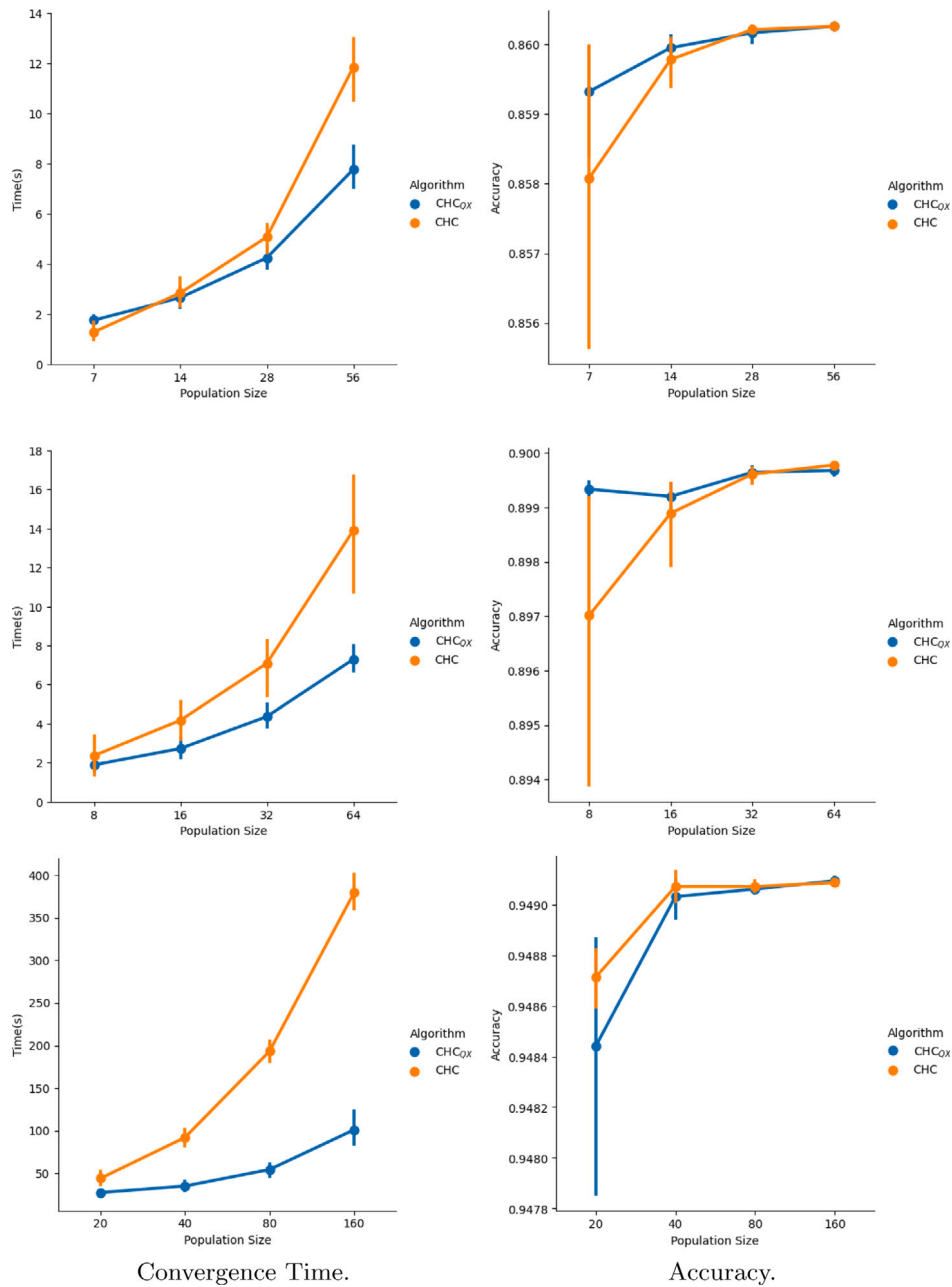

**Fig. 4.** The effect of varying population size on convergence time and accuracy for $\text{CHC}_{QX}$ and CHC. The results of datasets are presented in the row order: `adult` (top), `bank-full` (middle) and `census-income` (bottom).

The time complexity of $\text{CHC}_{QX}$ consists of the time required to construct the meta-model, and then the time of feature selection. The time complexity of one generation of $\text{CHC}_{QX}$ is variable, as the algorithm mostly uses the computationally cheap meta-model to carry fitness evaluations, and only uses the true fitness evaluation occasionally. The amortized cost of one generation of $\text{CHC}_{QX}$ is:

$$\frac{T_{CHC_{QX}}(r)}{r} = \frac{T_{is}(r) + T_{fs}(r)}{r}$$

where $T_{is}(r)$ is the run-time of the instance selection stage, and $T_{fs}(r)$ is the run-time of the feature selection stage.

The instance selection of $\text{CHC}_{QX}$ involves evaluating a fixed number of randomly selected feature subsets using the original function, the number of evaluations is a hyper-parameter of the algorithm denoted as $q$, with a default value of 10. The total time complexity of this operation is accordingly $q \cdot n \cdot k^2$.

$\text{CHC}_{QX}$ uses an instance selection GA to select samples to construct a meta-model that offers the best trade-off between highest correlation (agreement with original function), and the smallest sample of instances. The overall run-time of the instance selection GA is hard to predict due to the stochastic nature of GA. But we estimate the worst case time of this stage based on our selection of hyper-parameters. We used a population of 4 individuals and carried evolution for a maximum of 10 generations, we are also using a "no change" counter to early stop instance selection if fitness does not improve for a number of generations (we set this value to 3). The worst case total number of evaluations accordingly is 40. We initialize the starting population with individuals with no more than $\frac{n}{2}$ selected instances. As we used CHC to carry instance selection, the number of selected instances will never exceeds $\frac{n}{2}$ during evolution due to the averaging effect of the HUX crossover operator of CHC. We provide the proof in Lemma 1.

---

[7] The number of evaluations per generation is variable for CHC because of incest prevention. For CHC case, $e$ could represent the expected number of evaluations per generation instead.

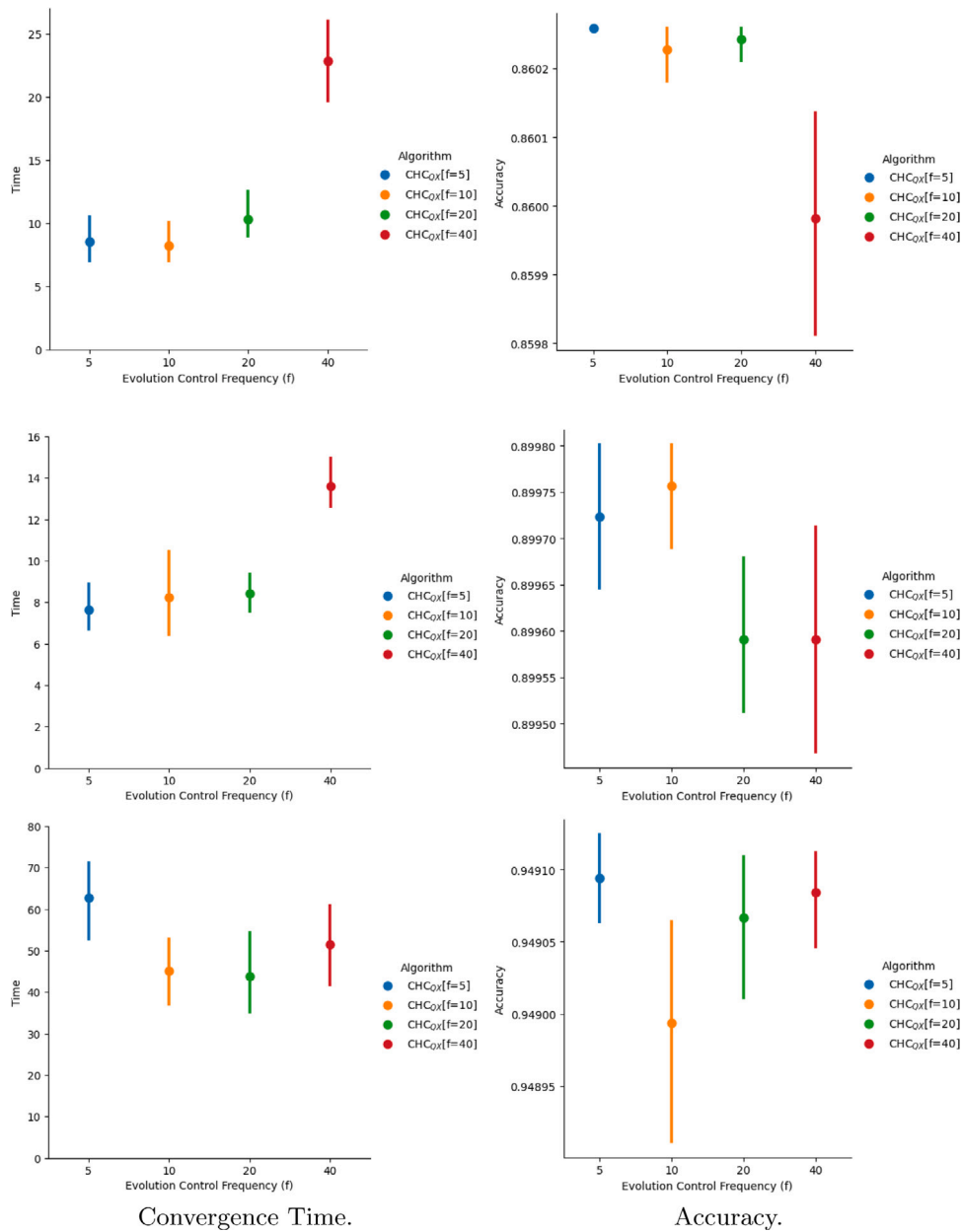

**Fig. 5.** The effect of varying evolution control frequency on convergence time and accuracy for $CHC_{QX}$. The results of datasets are presented in the row order: `adult` (top), `bank-full` (middle) and `census-income` (bottom).

**Lemma 1.** *The HUX crossover in binary optimization produces off-springs with the number of 1s equal to the average of their parents number of 1s.*

**Proof.** Given two binary strings individuals $g_1$ and $g_2$, expressed as $g_1, g_2 \in \{0,1\}^k$, where $k$ indicates the length of the string, we define $n_1, n_2$ as the total number of 1s in $g_1$ and $g_2$, respectively.

The HUX operator copies all bits matched in both parents, and then copies half of the different bits from each parent. The probability to have 1 in the same bit of $g_1, g_2$ is $(\frac{n_1}{k}.\frac{n_2}{k})$, while the probability to have different bits is $(\frac{n_1}{k}.(1-\frac{n_2}{k}) + \frac{n_2}{k}.(1-\frac{n_1}{k}))$. We can accordingly calculate $n'_1, n'_2$ for off-springs $g'_1, g'_2$ as follows:

$$\frac{n'_1}{k} = \frac{n'_2}{k} = \frac{n_1}{k}.\frac{n_2}{k} + \frac{1}{2}.\frac{n_1}{k}.(1-\frac{n_2}{k}) + \frac{1}{2}.\frac{n_2}{k}.(1-\frac{n_1}{k})$$

$$= \frac{n_1.n_2}{k^2} + \frac{n_1}{2.k} - \frac{n_1.n_2}{2.k^2} + \frac{n_2}{2.k} - \frac{n_2.n_1}{2.k^2}$$

$$= \frac{n_1.n_2}{k^2} + \frac{n_1 + n_2}{2.k} - \frac{n_1.n_2}{k^2}$$

$$= \frac{n_1 + n_2}{2.k}$$

This indicates that $n'_1 = n'_2 = \frac{n_1+n_2}{2}$

Based on Lemma 1, we can guarantee that the instance selection phase will never produce individuals larger than the maximum of the initial population (with $\frac{n}{2}$ instances). Accordingly, The worst case time for the instance selection phase of $CHC_{QX}$ is:

$$T_{is}(r) = 40 \cdot q \cdot \frac{n}{2} \cdot k^2 + q \cdot n \cdot k^2 = 21 \cdot q \cdot n \cdot k^2$$

During feature selection, $CHC_{QX}$ carries fitness evaluations mostly using the meta-model. Only on a predefined number of generations $f$ (by default, $f = 10$), the algorithm re-evaluates all individuals in the population using the original function. Therefore, the total time of carrying r generations of feature selection using $CHC_{QX}$ is:

$$T_{fs}(r) = r \cdot e \cdot n' \cdot k^2 + \left\lceil \frac{r}{f} \right\rceil \cdot e \cdot n \cdot k^2$$

The number of instances in the meta-model is in the worst case $n' = \frac{n}{2}$ based on Lemma 1. Therefore, the worst case total run of the feature selection phase is:

$$T_{fs}(r) = (\frac{r}{2} + \left\lceil \frac{r}{f} \right\rceil) \cdot e \cdot n \cdot k^2$$

The total cost of $r$ generations of $CHC_{QX}$ is:

$$T_{CHC_{QX}}(r) = T_{is}(r) + T_{fs}(r) = (21 \cdot q + (\frac{r}{2} + \left\lceil \frac{r}{f} \right\rceil) \cdot e) \cdot n \cdot k^2$$

By using the algorithm hyper-parameters values of ($q = 10, f = 10$) the amortized cost of one generation of $CHC_{QX}$ is:

$$\frac{T_{CHC_{QX}}(r)}{r} = \frac{(210 + (\frac{r}{2} + \left\lceil \frac{r}{10} \right\rceil) \cdot e) \cdot n \cdot k^2}{r}$$

Naturally, the amortized overhead cost of the instance selection stage represented by $\frac{T_{is}(r)}{r} = \frac{210 \cdot n \cdot k^2}{r}$ is inversely proportional to the total number of generations during feature selection. For problems that can be solved with a small number of generations, the overhead cost outweighs the benefit of the algorithm. For the default hyper-parameters of $CHC_{QX}$, and a population of 50 individuals ($e = 50$) we can show that for $r = 13$ generations:

$$T_{CHC_{QX}}(r = 13) = (210 + 6.5 \cdot 50 + 2 \cdot 50) \cdot n \cdot k^2 = 635 \cdot n \cdot k^2$$

$$T_{CHC}(r = 13) = 13 \cdot 50 \cdot n \cdot k^2 = 650 \cdot n \cdot k^2$$

$$\Rightarrow T_{CHC_{QX}}(r = 13) < T_{CHC}(r = 13)$$

## 6. Conclusion

In this paper, we have proposed a two-staged surrogate-assisted solution for the computational problem of using GA for feature selection by constructing a meta-model for fitness evaluations following a qualitative approximation approach. We defined the term "Approximation Usefulness" and used the expected value of rank correlation to quantify correctness of evolutionary selections, and quality of constructed meta-models.

According to our experiments, an Approximation Usefulness Curve follows an inverse power law function similar to the Learning Curve. In the left part of the curve, the quality of a meta-model improves rapidly with more training data, until it reaches a stage in which adding more data improves the quality of the meta-model very slowly, and eventually it stops altogether.

We carried an amortized analysis of the computation time of $CHC_{QX}$ and show that the amortized cost of one generation of our algorithm is in the worst case smaller than its counterpart algorithm CHC, as long as the two algorithms run for at least 13 generations. This analysis is supported by our empirical results where $CHC_{QX}$ demonstrated better scalability as datasets grow larger (in terms of number of instances). We further validated our findings by also creating a variant of the PSO algorithm $PSO_{QX}$ and demonstrated similar results using different meta-heuristic.

It must be noted that although $CHC_{QX}$ can be used with any learning algorithm, we have deliberately used Decision Tree as the baseline learning algorithm in all the experiments we performed. Given that Decision Tree induction is based on a greedy top-down approach of splitting the data based on an impurity measure, non-informative features will not be selected for the top nodes of the tree. The implicit feature selection of Decision Tree makes it a challenging choice for our experiments. We have observed in our results how feature selection using our meta-models consistently and significantly improved the baseline performance of Decision Tree. Our results confirm the superiority of the global search of GA in comparison to the local or greedy search of a traditional Decision Tree. Interested readers could refer to Barros, Basgalupp, De Carvalho, and Freitas (2011) to learn more about how an evolutionary approach might overcome the shortcomings of a greedy search in a traditional Decision Tree.

## 7. Limitations

The instance selection stage of $CHC_{QX}$ involves evaluating a fixed number of randomly generated feature subsets using the original function. Naturally, we would expect the instance selection to produce higher quality meta-models given a higher number of solutions evaluated using the original function. Clearly, the more evaluations we observe from the optimization surface of the original function, the easier it gets to produce a qualitatively similar meta-model. The obvious trade-off is the one between computational cost and quality of meta-model. As the main goal of this work is to make the evolutionary process time efficient ideally, we would prefer making the smallest number of evaluations using the expensive original function. Additionally, we have used a simple uniform approach with a fixed probability to control the variability in the number of selected features within the fixed solutions. However, some recent studies are realizing improved diversity and performance using low-discrepancy sequences (Bangyal, Hameed, Alosaimi, & Alyami, 2021). The impact of the initialization method on the final outcome of our algorithm could be investigated further.

We have learned from the amortized analysis that in the run-time of one generation of $CHC_{QX}$ is more computationally efficient than a classical wrapper. However, the instance selection phase of $CHC_{QX}$ is still computationally expensive. The reason is that it involves several of the original function evaluations. A better understanding of the basis of instance selection of $CHC_{QX}$ could lead us to redefine the instance selection fitness function to be more computationally efficient. As the process of active selection of instances in $CHC_{QX}$ is realized using a GA, it is challenging to explain why certain instances are selected by our algorithm. We think a future work could either: evaluate the characteristics of the selected instances (e.g., distance to decision boundary), or analyse statistical measures of the selected sample subset (e.g., Kolmogorov–Smirnov test or KL Divergence). Explanation on the basis of instance selection is not only useful for the insights. It could potentially obviate the need to perform evaluations using the original function.

We have never considered the class imbalance case as it is not the focus of this paper. In all experiments accuracy was used to evaluate the fitness of feature subsets. This metric could naturally be replaced with one that accounts for the imbalance case, e.g., average recall. We believe our instance selection method would still handle the imbalance case as the fitness function is designed to construct a meta-model which is aligned with a model trained using all instances. Instance subsets that neglect some minority class will lead to meta-models with poor correlation with the original function. If imbalance is an issue, a stratified sampling approach could ensure that randomly initialized instance solutions will not ignore the minority class.

## 8. Future work

In this paper, our method of constructing qualitative meta-models is specifically applied to the feature selection problem. An extension of this work would be to apply the same algorithm to other, related optimization problems, for example the hyper-parameters tuning of Machine Learning models using GA. The process of identifying the best hyper-parameter combinations shares many of the same computational challenges as the feature selection task. However, we should highlight that the fundamental notion that allows $CHC_{QX}$ to work for the feature selection problem might not hold for hyper-parameters tuning. A model trained with a small number of samples would rank different feature subsets similarly to a model trained with all available data. Intuitively, both models will struggle to improve generalization performance using "bad" features. This notion however, might not hold for the hyper-parameters tuning problem and needs to be investigated thoroughly. For example, intuitively, working with larger data sets permits the construction of more complex models without overfitting.

In the future, we also intend to study the differences in characteristics of the optimization surfaces between the original function and the meta-model. It is possible that this makes a difference; if, for example, the approximation is a lot less smooth than the original (or, on the contrary, very flat). It might turn out to be a lot more difficult for the optimization procedure — even if the maximum is correct, and it may be harder (or easier) to find it. However, we do not address this question within the scope of this work, only relying on Eqs. (1) and (2) to measure the quality of meta-models.

$PSO_{QX}$ demonstrated a capability to perform feature selection for data sets of high dimensional features (e.g., qsar with 1024 features). This capability makes $PSO_{QX}$ useful for Deep Learning applications with high dimensional input (e.g., images). A future work is planned to apply and evaluate $PSO_{QX}$ using such data sets and learning algorithms.

## CRediT authorship contribution statement

**Mohammed Ghaith Altarabichi:** Proposed the main idea of the work, Development of the work, Methodology, Experiments, Writing – original draft. **Sławomir Nowaczyk:** Development of the work, Methodology, Experiments, Writing – original draft. **Sepideh Pashami:** Development of the work, Methodology, Experiments, Writing – original draft. **Peyman Sheikholharam Mashhadi:** Development of the work, Methodology, Experiments, Writing – original draft.

## Declaration of competing interest

The authors declare that they have no known competing financial interests or personal relationships that could have appeared to influence the work reported in this paper.

## Data availability

All the data sets used in this paper are publicly available from the UCI ML repository.[8]

## Acknowledgment

This research has been funded by Vinnova, Strategic Vehicle Research and Innovation Programme.

---

[8] http://archive.ics.uci.edu/ml